%% file: main.tex
\documentclass[10pt,twocolumn,letterpaper]{article}

\usepackage{iccv}              %

\usepackage[ruled,linesnumbered]{algorithm2e}

\SetKwInput{KwData}{Input}
\SetKwInput{KwResult}{Output}
\SetKwBlock{Loop}{loop}{end}
\SetKwComment{Comment}{/* }{ */}
\SetAlFnt{\small}
\SetAlCapFnt{\small}
\SetAlCapNameFnt{\small}
\SetAlCapHSkip{0pt}

\input{packages}

\input{macros}

\definecolor{iccvblue}{rgb}{0.21,0.49,0.74}
\usepackage[pagebackref,breaklinks,colorlinks,allcolors=iccvblue]{hyperref}

\def\paperID{0000} %
\def\confName{ICCV}
\def\confYear{2025}

\title{Generating Physically Stable and Buildable Brick Structures from Text}

\author{Ava Pun\thanks{Indicates equal contribution.} \quad Kangle Deng\footnotemark[1] \quad Ruixuan Liu\footnotemark[1] \quad Deva Ramanan \quad Changliu Liu \quad Jun-Yan Zhu \\
Carnegie Mellon University
}

\begin{document}

\input{figTex/teaser}
\input{sections/00_abstract}

\input{sections/01_intro}
\input{sections/02_related_work}
\input{sections/03a_dataset}
\input{sections/03b_method}

\input{sections/04_experiment}

\input{sections/05_discussion}

\clearpage

\section*{Acknowledgments}
We thank Minchen Li, Ken Goldberg, Nupur Kumari, Ruihan Gao, and Yihao Shi for their discussions and help.  
We also thank Jiaoyang Li, Philip Huang, and Shobhit Aggarwal for developing the bimanual robotic system.
This work is partly supported by the Packard Foundation, Cisco Research Grant, and Amazon Faculty Award. This work is also in part supported by the Manufacturing Futures Institute,
Carnegie Mellon University, through a grant from the Richard King Mellon
Foundation. KD is supported by the Microsoft Research PhD Fellowship.

{
    \small
    \bibliographystyle{ieeenat_fullname}
    \bibliography{main}
}

\input{06_supplement}

\end{document}

%% file: packages.tex
\usepackage[accsupp]{axessibility}  %

\usepackage[T1]{fontenc}
\usepackage{times}
\usepackage{epsfig}
\usepackage{graphicx}
\usepackage{amsmath}
\usepackage{amssymb}
\usepackage{color,xcolor}

\usepackage{comment}

\usepackage{pifont}

\usepackage{microtype}
\usepackage[font=small]{caption}
\usepackage{arydshln}
\usepackage{tabularx}
\usepackage{adjustbox}
\usepackage{array}
\usepackage{booktabs}
\usepackage{colortbl}
\usepackage{float,wrapfig}
\usepackage{hhline}
\usepackage{multirow}
\usepackage{subcaption} %
\usepackage[percent]{overpic}

\usepackage{stmaryrd}

\usepackage{amsmath,amsfonts,amssymb}
\usepackage{bm}
\usepackage{nicefrac}
\usepackage{microtype}
\usepackage{dsfont}
\usepackage{changepage}
\usepackage{extramarks}
\usepackage{fancyhdr}
\usepackage{setspace}
\usepackage{soul}
\usepackage{xspace}

\usepackage{enumitem}
\usepackage[title]{appendix}

\usepackage{duckuments}
\usepackage{subcaption}

\usepackage{animate}
\usepackage{tabularx}

\usepackage[multiple]{footmisc}

\usepackage{amsmath}
\usepackage{amssymb}
\usepackage{booktabs}
\usepackage{cuted}
\usepackage{capt-of}
\usepackage{graphicx}
\usepackage{xcolor}
\usepackage{comment}
\usepackage{multirow}
\usepackage{bm}

\usepackage[font=small]{caption}
\usepackage{color,xcolor}
\usepackage{epsfig}
\usepackage{comment}
\usepackage{pifont}
\usepackage{microtype}
\usepackage{arydshln}
\usepackage{tabularx}
\usepackage{adjustbox}
\usepackage{array}
\usepackage{colortbl}
\usepackage{float,wrapfig}
\usepackage{hhline}
\usepackage{multirow}
\usepackage[percent]{overpic}
\usepackage{stmaryrd}
\usepackage{breqn}
\usepackage{bm}
\usepackage{nicefrac}
\usepackage{dsfont}
\usepackage{changepage}
\usepackage{extramarks}
\usepackage{fancyhdr}
\usepackage{soul}
\usepackage{xspace}
\usepackage{enumitem}
\usepackage[title]{appendix}
\usepackage{balance}
\usepackage{booktabs}
\usepackage{siunitx}
\usepackage{subcaption}

\usepackage{natbib}

%% file: macros.tex
\definecolor{MyPurple}{RGB}{111,0,255}

\definecolor{MyTeal}{RGB}{0,179,185}

\newcommand{\h}{h}

\newcommand{\reffig}[1]{Figure~\ref{fig:#1}}
\newcommand{\refsec}[1]{Section~\ref{sec:#1}}
\newcommand{\refapp}[1]{Appendix~\ref{sec:#1}}
\newcommand{\reftbl}[1]{Table~\ref{tbl:#1}}
\newcommand{\refalg}[1]{Algorithm~\ref{alg:#1}}

\newcommand{\refeq}[1]{Eqn.~\ref{eq:#1}}

\newcommand{\lblfig}[1]{\label{fig:#1}}
\newcommand{\lblsec}[1]{\label{sec:#1}}
\newcommand{\lbleq}[1]{\label{eq:#1}}

\newcommand{\lblalg}[1]{\label{alg:#1}}

\newcommand{\ignorethis}[1]{}

\newcommand{\myparagraph}[1]{\vspace{3pt} \noindent \textbf{#1} \ } %

\def\1{\bm{1}}

\DeclareMathOperator*{\argmin}{arg\,min}

\newcolumntype{L}[1]{>{\raggedright\let\newline\\\arraybackslash\hspace{0pt}}m{#1}}
\newcolumntype{C}[1]{>{\centering\let\newline\\\arraybackslash\hspace{0pt}}m{#1}}
\newcolumntype{R}[1]{>{\raggedleft\let\newline\\\arraybackslash\hspace{0pt}}m{#1}}

\newcommand{\ignore}[1]{}

\makeatletter
\DeclareRobustCommand\onedot{\futurelet\@let@token\@onedot}
\def\@onedot{\ifx\@let@token.\else.\null\fi\xspace}
\def\eg{e.g\onedot,\xspace} 

\def\ie{i.e\onedot,\xspace}

\makeatother

\newcommand{\LEGOtm}{LEGO\texorpdfstring{\textsuperscript{\textregistered}}{}}
\newcommand{\SLV}{StableText2Brick}
\newcommand{\LegoGPT}{\textsc{BrickGPT}}
\newcommand{\model}{\theta\xspace}
\newcommand{\mesh}{\mathcal{M}\xspace}
\newcommand{\voxelgrid}{\mathcal{V}\xspace}
\newcommand{\texturecaption}{c\xspace}

\newcommand{\legostructure}{B=[b_1, b_2, \dots, b_N]\xspace}

%% file: figTex/teaser.tex
\maketitle

\begin{strip}\centering
\vspace{-45pt}
\includegraphics[width=\linewidth]{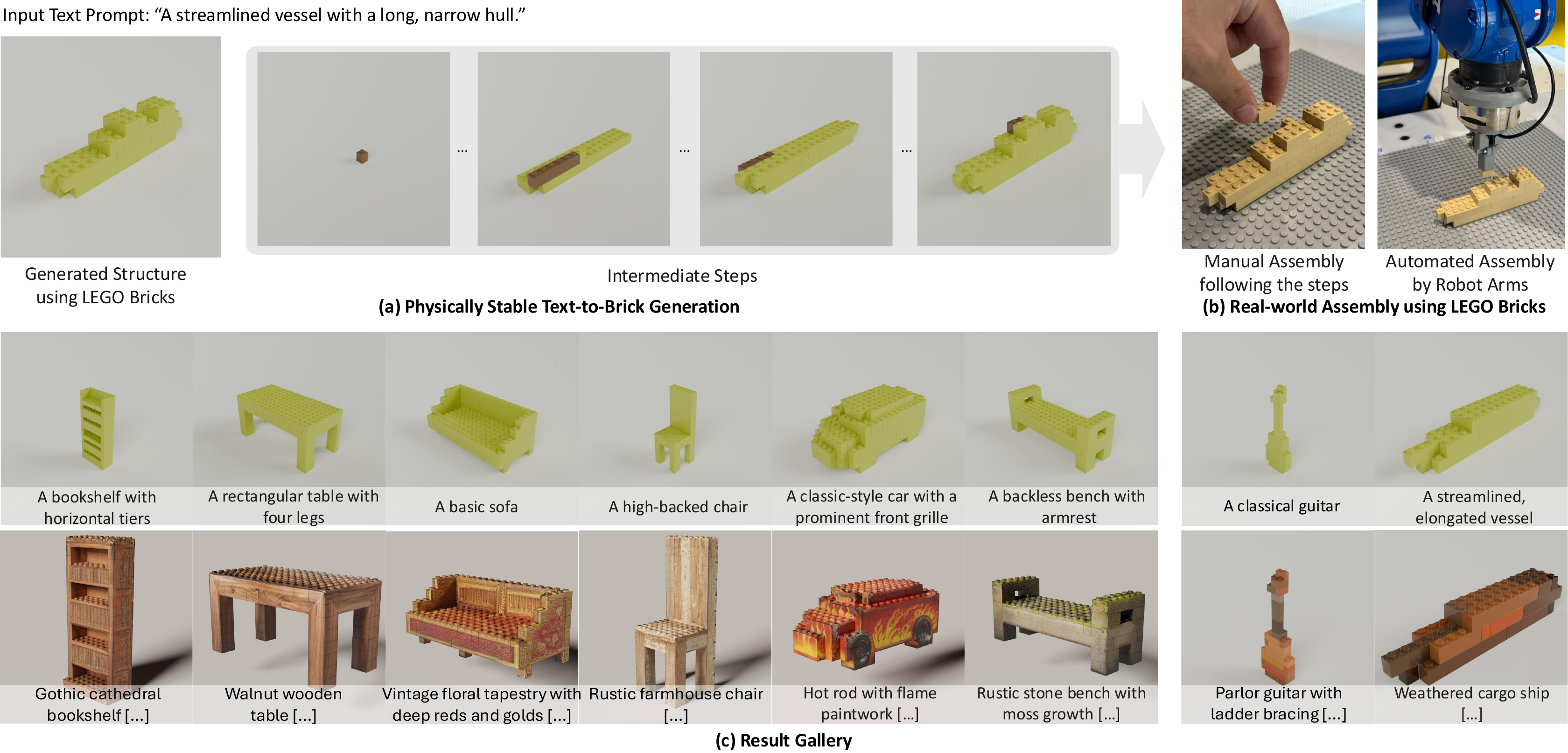}
\captionof{figure}{\textbf{Overview of \LegoGPT{}.} (a) Our method generates physically stable interconnecting brick assembly structures from text descriptions through an end-to-end approach, showing intermediate brick-by-brick steps. (b) The generated designs are buildable both by hand and by automated robots. (c) We show example results with corresponding text prompts. Besides basic brick designs (top), our method can generate colored (bottom right) and textured (bottom left) models with appearance descriptions. We highly recommend the reader to check our \href{https://avalovelace1.github.io/BrickGPT/}{website} for step-by-step videos.
\lblfig{teaser}
}
\vspace{-5pt}
\end{strip}

%% file: sections/00_abstract.tex
\begin{abstract}
    We introduce \LegoGPT{}, the first approach for generating physically stable interconnecting brick assembly models from text prompts. To achieve this, we construct a large-scale, physically stable dataset 
    of brick structures, along with their associated captions, and train an autoregressive large language model to predict the next brick to add via next-token prediction. To improve the stability of the resulting designs, we employ an efficient validity check and physics-aware rollback during autoregressive inference, which prunes infeasible token predictions using physics laws and assembly constraints. Our experiments show that \LegoGPT{} produces stable, diverse, and aesthetically pleasing brick structures that align closely with the input text prompts. We also develop a text-based brick texturing method to generate colored and textured designs. We show that our designs can be assembled manually by humans and automatically by robotic arms. We release our new dataset, \SLV{}, containing over 47,000 brick structures of over 28,000 unique 3D objects accompanied by detailed captions, along with our code and models at the project website: {\small\url{https://avalovelace1.github.io/BrickGPT/}}.
    
\end{abstract}

%% file: sections/01_intro.tex
\section{Introduction}

3D generative models have made remarkable progress, driven by advances in generative modeling~\cite{goodfellow2014generative,sohl2015deep} and neural rendering~\cite{kerbl3Dgaussians,mildenhall2020nerf}. These models have enabled various applications in virtual reality, gaming, entertainment, and scientific computing. Several works have explored synthesizing 3D objects from text~\cite{poole2022dreamfusion}, adding texture to meshes~\cite{richardson2023texture,deng2024flashtex}, and manipulating the shape and appearance of existing 3D objects and scenes~\cite{liu2021editing,instructnerf2023}.

However, creating real-world objects with existing methods remains challenging. Most approaches focus on generating diverse 3D objects with high-fidelity geometry and appearance~\cite{zhang2024clay,hong2023lrm}, but these digital designs often cannot be physically realized due to two key challenges~\cite{makatura2023can}. 
First, the objects may be difficult to assemble using standard components. Second, the resulting structure may be physically unstable even if assembly is possible. Without proper support, parts of the design could collapse, float, or be disconnected. %

In this work, we address the challenge of generating \emph{physically realizable objects}. We study this problem in the context of designing structures made of interlocking toy bricks, such as \LEGOtm{} blocks. These are widely used in education, artistic creation, and manufacturing prototyping. Additionally, they can serve as a reproducible benchmark, as all standard components are readily available. 
Due to the significant effort required to design brick structures manually, recent studies have developed automated algorithms to streamline the process and generate compelling results. However, existing approaches primarily create structures from a given 3D object~\cite{legolization} or focus on a single object category~\cite{geLearnCreateSimple2024,geCreatingLEGOFigurines2024}. 

Our goal is to generate brick assembly structures directly from freeform text prompts while ensuring physical stability and buildability. Specifically, we aim to train a generative model that produces designs that are:
\begin{itemize}
   \item \emph{Physically stable}: 
  Built on a baseplate with strong structural integrity, without floating or collapsing bricks. 
    \item \emph{Buildable}: Compatible with standard toy brick pieces and able to be assembled brick-by-brick by humans or robots. 
\end{itemize}

In this work, we introduce \LegoGPT{} with the key insight of repurposing autoregressive large language models, originally trained for next-token prediction, for next-brick prediction.  We formulate the problem of brick structure design as an autoregressive text generation task, where the next-brick dimension and placement are specified with a simple textual format. To ensure generated structures are both {\em stable} and {\em buildable}, we enforce physics-aware assembly constraints during both training and inference. During training, we construct a large-scale dataset of physically stable brick structures paired with captions. 
During autoregressive inference, we enforce feasibility with an efficient validity check and physics-aware rollback to ensure that the final tokens adhere to physics laws and assembly constraints.

Our experiments show that the generated designs are stable, diverse, and visually appealing while adhering to input prompts. Our method outperforms pre-trained LLMs with or without in-context learning, and previous approaches based on 3D mesh generation. 
Finally, we explore applications such as text-driven brick texturing, as well as manual assembly and automated robotic assembly of our designs. Our dataset, code, and models are available at the project website: {\small\url{https://avalovelace1.github.io/BrickGPT/}}. 

%% file: sections/02_related_work.tex
\section{Related Work}

\myparagraph{Text-to-3D Generation.}
Text-to-3D generation has seen remarkable progress recently, driven by advances in neural rendering and generative models. Dreamfusion~\cite{poole2022dreamfusion} and Score Jacobian Chaining~\cite{wang2023sjc} pioneer zero-shot text-to-3D generation by optimizing neural radiance fields~\cite{mildenhall2020nerf} with pre-trained diffusion models~\cite{rombach2022high}. 
Subsequent work has explored alternative 3D representations~\cite{lin2023magic3d, Chen_2023fantasia3D, metzer2022latent, sun2023dreamcraft3d, sweetdreamer, long2023wonder3d, michel2022text2mesh} and improved loss functions~\cite{wang2023prolificdreamer, ye2025dreamreward, mcallister2024rethinking, tran2025diverse, katzir2023noise, lukoianov2024sdi}. %
Rather than relying on iterative optimization, a promising alternative direction trains generative models directly on 3D asset datasets, with various backbones including diffusion models~\cite{shue2023triplane_diffusion, zhou2021point_voxel_diffusion, li2023diffusion_sdf, nam20223d_ldm, zhang2024clay,li2024craftsman,hong20243dtopia,zhao2025hunyuan3d20scalingdiffusion,XCube}, large reconstruction models~\cite{hong2023lrm,li2023instant3d,xu2023dmv3d,TripoSR2024}, U-Nets~\cite{lgm,liu2025meshformer}, and autoregressive models~\cite{polygen, meshxl, edgerunner, meshgpt, meshanything, chen2024meshanythingv2, hao2024meshtron, weng2024scaling}. %
However, these existing methods cannot be directly applied to generating brick structures as they do not account for the unique physical and assembly constraints of real-world designs~\cite{makatura2023can}. We bridge this gap by introducing a method for generating physically stable and buildable brick structures directly from text prompts. 

\myparagraph{Autoregressive 3D Modeling.} 
Recent research has successfully used autoregressive models to generate 3D meshes~\cite{polygen, meshgpt, meshxl, edgerunner, meshanything, chen2024meshanythingv2, hao2024meshtron, weng2024scaling,deng2025efficient}, often conditioned on input text or images.  
Most recently, LLaMA-Mesh~\cite{llamamesh} demonstrates that large language models (LLMs) can be fine-tuned to output 3D shapes in plain-text format, given a text prompt. However, most existing autoregressive methods focus on mesh generation.
In contrast, we focus on generating brick structures from text prompts, leveraging LLMs' reasoning capabilities.

\input{figTex/dataset}

\myparagraph{Brick Assembly and Design Generation.}
Creating brick structures given a reference 3D shape has been widely studied \cite{Kim2014SurveyOA}.
Existing works~\cite{LEGOBuilder,testuz2013automatic,automaticgeneration} formulate the generation as an optimization problem guided by hand-crafted heuristics, such as ensuring that all bricks are interconnected and minimizing the number of bricks.
\citet{wang2022translating} translate a visual manual into step-by-step brick assembly instructions.
\citet{legolization} use stability analysis to find weak structural parts and rearrange the local brick layout to generate physically stable designs.
\citet{kim2020combinatorial,liu2024physics} formulate a planning problem to fill the target 3D model sequentially. 
However, these methods only generate designs given an input 3D shape, assuming a valid brick structure exists, which is difficult to verify in practice.

Few works have explored learning-based techniques to generate toy brick designs.  \citet{thompsonBuildingLEGOUsing2020} use a deep graph generative model in which the graph encodes brick connectivity. %
However, this method is limited to generating simple classes %
using a single brick type. More recently, \citet{geLearnCreateSimple2024} use a diffusion model to predict a semantic volume, which is then translated into a high-quality micro building. %
Their method produces impressive results for a single category. %
\citet{zhouBrickYourself32022} and \citet{geCreatingLEGOFigurines2024} generate compelling figurine designs given an input portrait, selecting components from a pre-made set that best match an input photo. Although effective for faces, extending this selection-based approach to arbitrary objects is challenging. \citet{legosketchart} formulate an optimization problem to create a brick model from an input image. While their output is a 2D brick mosaic, we focus on 3D structures in this work.
\citet{goldberg2025bloxnet} query a vision-language model to generate diverse 3D assembly structures. However, they use regular building blocks instead of bricks with interlocking connections, and thus the structures have limited expressiveness. 
\citet{lennonImage2LegoCustomizedLEGO2021} voxelize an input image and use heuristics to convert the voxels into a brick model without considering physical constraints. In contrast, our method performs the text-to-brick task without requiring intermediate voxel representations.

\myparagraph{Physics-Aware Generation.} 
Physics-aware 3D generation can be broadly categorized into two approaches: direct constraint enforcement and learned validation. Simple physical constraints, such as collision avoidance and contact requirements, can be incorporated directly through explicit penalty terms during optimization~\cite{misra2025shapeshift, yang2024physcene, yuan2023physdiff, vilesov2023cg3d, liu2023few, huang2023diffusionscenes,guo2024physcomp}. More complex physical properties, such as structural stability and dynamic behavior, typically require physics simulators~\cite{evobodybuild,ni2024phyrecon,xu2024precise,mezghanni2022physical,goldberg2025bloxnet} or data-driven physics-aware assessment models~\cite{dong2024gpld3d,mezghanni2021physically}.
To our knowledge, our paper is the first attempt to incorporate physics-aware constraints into text-based brick assembly structure generation.

%% file: figTex/dataset.tex
\begin{figure*}
    \centering
    \includegraphics[width=0.95\linewidth]{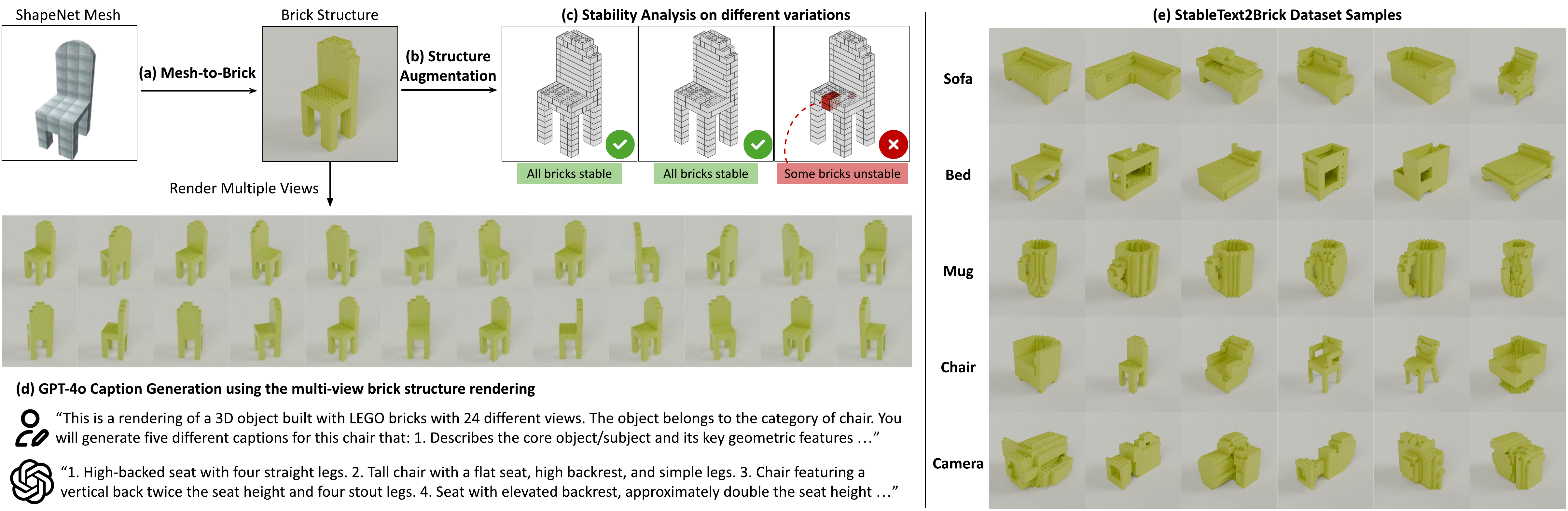}
    \caption{\textbf{\SLV{} Dataset.} (a) From a ShapeNetCore~\cite{shapenet2015} mesh, we generate a brick structure by voxelizing it onto a $20\times 20 \times 20$ grid, then constructing its brick layout with a delete-and-rebuild algorithm. (b) We augment each shape with multiple structural variations by randomizing the brick layout while preserving the overall shape. (c) Stability analysis~\cite{liuStableLegoStabilityAnalysis2024} is performed on each variation to filter out physically unstable designs. (d) To obtain captions for each shape, we render the brick structure from 24 different viewpoints and use GPT-4o~\cite{achiam2023gpt} to generate detailed geometric descriptions. (e) Data samples from 5 categories in our \SLV{} dataset.}
    \lblfig{dataset}
    \vspace{-7pt}
\end{figure*}

%% file: sections/03a_dataset.tex
\section{Dataset}
\lblsec{dataset}

Training a modern autoregressive model requires a large-scale dataset. Therefore, we introduce \SLV{}, a new dataset that contains $47{,}000+$ different toy brick assembly structures, covering $28{,}000+$ unique 3D objects from $21$ common object categories of the ShapeNetCore dataset~\cite{shapenet2015}. 
We select categories featuring diverse and distinctive 3D objects while excluding those resembling cuboids. %
Each structure is paired with a group of text descriptions and a stability score, which indicates its physical stability and buildability. Below, we describe the dataset construction, an overview of which is given in \reffig{dataset}.

\input{figTex/overview}

\myparagraph{Brick Representation.} 
We consider brick structures built on a fixed baseplate.
Each structure in \SLV{} is represented as $\legostructure$ with $N$ bricks, and each element denotes a brick's state as $b_i=[h_i, w_i, x_i, y_i, z_i]$. 
Here, $h_i$ and $w_i$ indicate the brick length in the $X$ and $Y$ directions, respectively, and $x_i$, $y_i$, and $z_i$ denote the position of the stud closest to the origin.
The position has $x_i\in [0,1,\dots,H-1]$, $y_i\in[0,1,\dots,W-1]$, $z_i\in[0,1,\dots,D-1]$, where $H$, $W$, and $D$ represent the dimensions of the discretized grid world.

\myparagraph{Shape-to-Brick.} We construct the dataset by converting 3D shapes from ShapeNetCore~\cite{shapenet2015} into brick structures as shown in \reffig{dataset}(a). 
Given a 3D mesh, we voxelize and downsample it into a $20\times 20 \times 20$ grid world to ensure a consistent scale, \ie $H=W=D=20$.
The brick layout is generated by a delete-and-rebuild algorithm similar to~\cite{legolization}.
To improve data quality and diversity, we introduce randomness and generate multiple different structures for the same 3D object, as illustrated in \reffig{dataset}(b).
This increases the chance of obtaining a stable structure and more diverse layouts.
More details can be found in \refapp{sec:training-details}.

\myparagraph{Stability Score.} 
We assess the physical stability of each structure, as illustrated in \reffig{dataset}(c), using the analysis method~\cite{liuStableLegoStabilityAnalysis2024}.
For a structure $\legostructure$, the stability score $S\in \mathbb{R}^{N}$ assigns each brick $b_i$ a value $s_i\in[0, 1]$ that quantifies the internal stress at its connections. Higher scores $s_i$ indicate greater stability, while $s_i=0$ indicates an unstable brick that will cause structural failure. 
Calculating the stability score requires solving a nonlinear program to determine the forces acting on each brick to achieve a static equilibrium that prevents structural collapse, as detailed in \refsec{physical_stability}. 
For typically-sized (\ie $<200$ bricks) structures in \reffig{dataset}, stability analysis takes $\sim$0.35 seconds on average. 
A structure is stable if all bricks have stability scores greater than 0; we only include stable structures in the \SLV{} dataset.

\myparagraph{Caption Generation.}
To obtain captions for each shape, we render the brick structure from 24 different viewpoints and combine them into a single multi-view image. 
We then prompt GPT-4o~\cite{achiam2023gpt} to produce five descriptions for these renderings with various levels of detail. Importantly, we ask GPT-4o to omit color information and focus only on geometry. %
The complete prompt is provided in \refapp{sec:training-details}.

\reffig{dataset}(e) shows several data samples from \SLV{}. 
The rich variations within each category and the comprehensive text-brick pairs make it possible to train large-scale generative models.
More insights on \SLV{} are discussed in \refapp{sec:training-details}.

%% file: figTex/overview.tex
\begin{figure*}[t]
    \centering
    \includegraphics[width=0.95\linewidth]{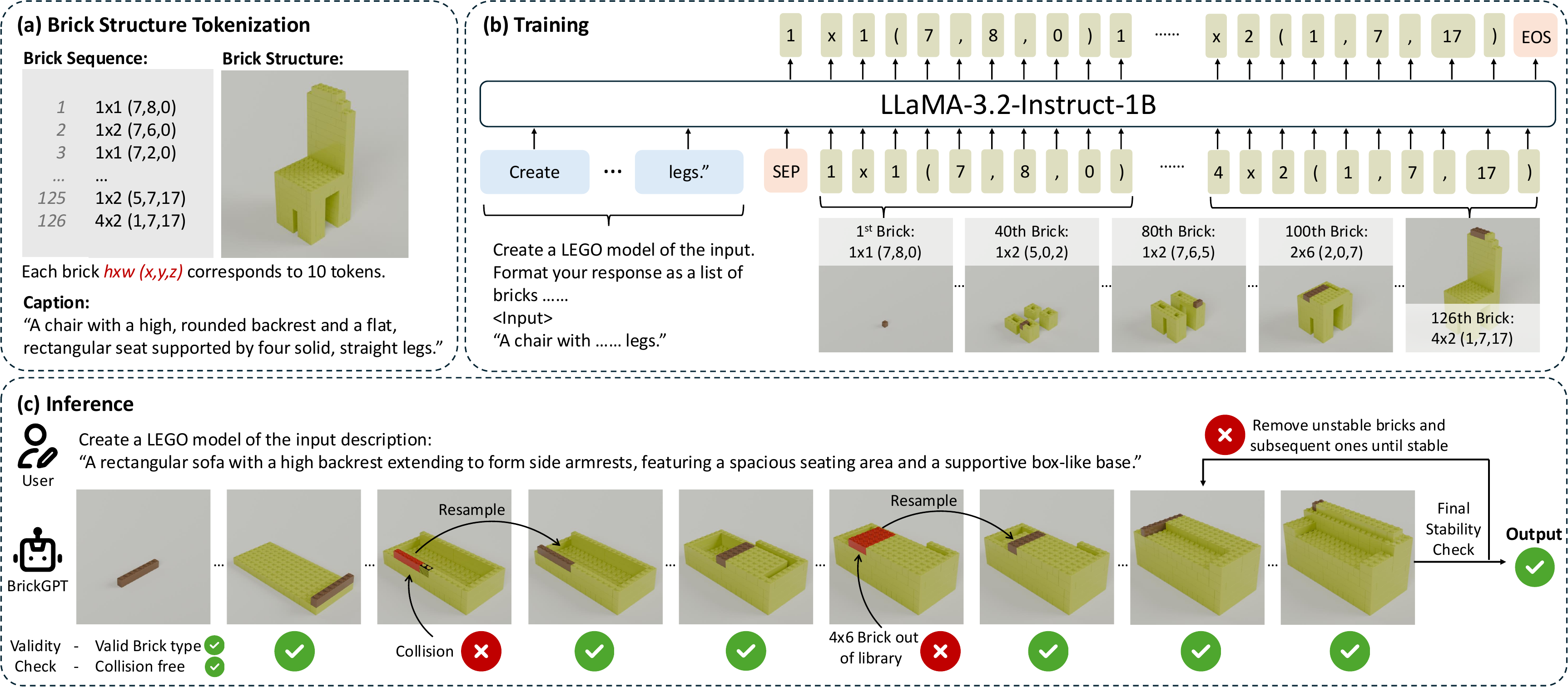}
    \caption{\textbf{Method.} (a) Our system tokenizes a brick structure into a sequence of text tokens, ordered in a raster-scan manner from bottom to top. %
    (b) We create an instruction dataset pairing brick sequences with descriptions to fine-tune LLaMA-3.2-Instruct-1B. (c) At inference time, \LegoGPT{} generates brick structures incrementally by predicting one brick at a time given a text prompt. For each generated brick, we perform validity checks to ensure it is well-formatted, exists in our brick library, and does not collide with existing bricks. After completing the design, we verify its physical stability. If the structure is unstable, we roll back to a stable state by removing all unstable bricks and their subsequent bricks, and resume generation from that point.}
    \lblfig{overview}
    \vspace{-7pt}
\end{figure*}

%% file: sections/03b_method.tex
\section{Method}

Here, we introduce \LegoGPT{}, a method for generating physically stable interconnecting toy brick assembly structures from text prompts. %
Leveraging LLMs' ability to model sequences and understand text, 
we fine-tune a pre-trained LLM for the brick structure generation task (\refsec{model-fine-tuning}). To increase the stability and buildability of our designs, we use brick-by-brick rejection sampling and physics-aware rollback during inference (\refsec{physical_stability}). \reffig{overview} illustrates an overview of our method.

\subsection{Model Fine-tuning}\lblsec{model-fine-tuning}

Pre-trained LLMs excel at modeling sequences and understanding natural language, making them promising candidates for our task. We further fine-tune a pre-trained LLM on a custom instruction dataset containing text prompts and their corresponding brick structures from \SLV{}.

\myparagraph{Pre-trained Base Model.} We use LLaMA-3.2-1B-Instruct~\cite{dubey2024llama} as our base model. %
This model is fine-tuned to give coherent answers to instruction prompts, making it suitable for text-based brick structure generation. 
As shown in \reffig{results}, the base model can generate brick structures through in-context learning, highlighting the promise of using pre-trained LLMs for our task. However, the generated structures fail to follow the input prompt, and they contain intersecting or disconnected bricks, making them physically unstable and unbuildable. To address these issues, we further fine-tune the pre-trained model using our \SLV{}. %

\myparagraph{Instruction Fine-tuning Dataset.} For each stable structure and its corresponding captions, we construct an instruction in the following format: ``(user) Create a LEGO model of \{caption\}. (assistant) \{brick-structure\}.'' To simplify training and reuse LLaMA's tokenizer, we represent brick structures in plain text. But what format should we use? The standard format LDraw~\cite{ldraw} has two main drawbacks. First, it does not directly include brick dimensions, which are crucial for assessing the structure and validating brick placements. Second, it includes unnecessary information such as arbitrary brick orientation and scale, which are redundant since all bricks are axis-aligned in our setting. %

Instead of using LDraw, we introduce a custom format to represent each brick structure. Each line of our format represents one brick as ``\{$h$\}$\times$\{$w$\} (\{$x$\},\{$y$\},\{$z$\})'', where $h \times w$ are brick dimensions and $(x,y,z)$ are its coordinates. All bricks are 1-unit-tall, axis-aligned cuboids, and the order of $\h$ and $w$ encodes the brick's orientation about the vertical axis. This format significantly reduces the number of tokens required to represent a design, while including brick dimension information essential for 3D reasoning. %
Bricks are ordered in a raster-scan manner from bottom to top.

With our fine-tuned \LegoGPT{} model $\theta$, we predict the bricks $b_1, b_2, ..., b_N$ in an autoregressive manner: 
\begin{align}\lbleq{autoregressive}
p(b_1, b_2, ..., b_N  | \model) = \prod_{i=1}^{N} p(b_i | b_1,...,b_{i-1}, \model) .
\end{align}

\subsection{Integrating Physical Stability}\lblsec{physical_stability}

Although trained on physically stable data, our model sometimes generates designs that violate physics and assembly constraints. To address this issue, we further incorporate physical stability verification into autoregressive inference.

\input{figTex/force_illustration}

A brick structure is considered physically stable and buildable if it does not collapse when built on a baseplate.
To this end, we assess physical structural stability using the stability analysis method~\cite{liuStableLegoStabilityAnalysis2024}.
We briefly overview this method below.
\reffig{force_model}(a) illustrates all possible forces exerted on a single brick.
We extend the single brick model and derive the structural force model $\mathcal{F}$, which consists of a set of candidate forces
as shown in \reffig{force_model}(b).
For a brick structure $\legostructure$, each brick $b_i$ has $M_i$ candidate forces $F_i^j\in \mathcal{F}_i, j\in[1, M_i]$.
A structure is stable if all bricks can reach static equilibrium, \ie
\begin{align}\lbleq{eq:equilibrium}
    \sum_j^{M_i}F_i^j=0, \qquad
    \sum_j^{M_i}{\tau}_i^j \dot= \sum_j^{M_i}L_i^j\times F_i^j=0,
\end{align}
where $L_i^j$ denotes the force lever corresponding to $F_i^j$.
The stability analysis is formulated into a nonlinear program as
\begin{equation}\lbleq{eq:obj}
\begin{split}
    \argmin_{\mathcal{F}}&\sum_i^N\Biggl\{\bigg|\sum_j^{M_i}F_i^j\bigg|+\bigg|\sum_j^{M_i}{\tau}_i^j\bigg|+\alpha \mathcal{D}_i^{\max}+\beta \sum \mathcal{D}_i\Biggl\},
    \end{split}
\end{equation}
subject to three constraints: 1) all force candidates in $\mathcal{F}$ should take non-negative values; 2) certain forces exerted on the same brick cannot coexist, \eg the pulling (red arrow) and pressing (blue arrow), the dragging (green arrow) and supporting (purple arrow); 3) Newton's third law, \eg at a given connection point, the supporting force on the upper brick should be equal to the pressing force on the bottom brick.
$\mathcal{D}_i\subset \mathcal{F}_i$ is the set of candidate dragging forces (green arrow) on $b_i$.
$\alpha$ and $\beta$ are hyperparameter weights.

Solving the above nonlinear program in \refeq{eq:obj} using Gurobi~\cite{gurobi} finds a force distribution $\mathcal{F}$ that drives the structure to static equilibrium with the minimum required internal stress, suppressing the overall friction (\ie $\sum \mathcal{D}_i$) as well as avoiding extreme values (\ie $ \mathcal{D}_i^{\max}$).
From the force distribution $\mathcal{F}$, we obtain the per-brick stability score as 
\begin{equation}\lbleq{eq:stability_score}
    \begin{split}
        s_i=\begin{cases}
            0 & %
            \begin{array}{c}
                  \sum_j^{M_i}F_i^j \ne 0 \\
                 \vee \quad \sum_j^{M_i}{\tau}_i^j \ne 0 \\
                 \vee \quad \mathcal{D}_i^{\max}>F_T,
            \end{array}
            \\
            \frac{F_T - \mathcal{D}_i^{\max}}{F_T} & \text{otherwise},
        \end{cases}
    \end{split}
\end{equation}
where $F_T$ is a measured constant friction capacity between brick connections.
Higher scores $s_i$ indicate greater stability, while $s_i=0$ indicates an unstable brick that will cause structural failure: either $\mathcal{F}$ cannot reach static equilibrium ($\sum_j^{M_i}F_i^j \ne 0 \vee \sum_j^{M_i}\tau_i^j \ne 0$) or the required friction exceeds the friction capacity of the material ($\mathcal{D}_i^{\max}>F_T$). 
Due to the equality constraints imposed by Newton's third law, \refeq{eq:obj} includes only the dragging forces and excludes pulling forces.
For a physically stable structure, we need $s_i>0, \forall i\in[1, N]$.

\myparagraph{When to apply stability analysis?}
A straightforward approach to ensuring physical stability is to apply stability analysis after each generated brick and resample a brick that would cause a collapse. 
However, this
step-by-step validation
could be time-consuming.
More importantly, many structures are unstable during construction yet become stable when fully assembled; adding a stability check after each brick generation could overly constrain the model exploration space. %
Instead, we propose brick-by-brick rejection sampling combined with physics-aware rollback to balance stability and diversity. %

\myparagraph{Brick-by-Brick Rejection Sampling.}
To improve inference speed and avoid overly constraining generation, we relax our constraints during inference.
First, when the model generates a brick and its position, the brick should be well-formatted (\eg available in the inventory) and lie within the world space. 
Second, the brick should not collide with the existing structure. Formally, for each generated brick $b_t$, we have $\mathcal{V}_t\cap \mathcal{V}_i = \varnothing, \forall i\in[1, t-1]$, where $\mathcal{V}_i$ denotes the voxels occupied by $b_i$.
These heuristics allow us to efficiently generate well-formatted structures without explicitly considering complex physical stability.
To integrate these heuristics, we use rejection sampling: if a brick violates the rules, we resample a new brick from the model. 
Due to the relaxed constraints, most bricks are valid, and rejection sampling does not significantly affect inference time.

\myparagraph{Physics-Aware Rollback.}
To ensure that the final design $\legostructure$ is physically stable, we calculate the stability score $S$.
If the resulting design is unstable, \ie $s_i=0, i\in\mathcal{I}$, we roll back the design to the state before the first unstable brick was generated, \ie $B'=[b_1, b_2, \dots, b_{\min{\mathcal{I}}-1}]$. Here, $\mathcal{I}$ is the set of the indices of all the unstable bricks. We repeat this process iteratively until we reach a stable structure $B'$, and continue generation from the partial structure $B'$. Note that we can use the per-brick stability score to efficiently find the collapsing bricks and their corresponding indices in the sequence.
We summarize our inference sampling in \refalg{alg:inference}.

\begin{algorithm}[t]
\footnotesize 
\caption{\footnotesize \LegoGPT{} inference algorithm. \lblalg{alg:inference}}
\KwData{Text prompt $\texturecaption$; Autoregressive model $\model$}
\KwResult{Brick structure following the text prompt}
$B \gets \text{empty brick structure}$\;
\Loop{
    \Comment{\scriptsize Predict next brick w/ rejection sampling}
    \For{$k=1,\ldots,\texttt{max_rejections}$}{
        $\texttt{context} \gets T \oplus B.\texttt{to_text_format}()$\;
        $b \gets \model.\texttt{predict_tokens}(\texttt{context})$ (\refeq{autoregressive})\;
        \lIf{$b$ is valid}{\textbf{break}} 
    }
    $B.\texttt{add_brick}(b)$\;
    \If(\tcp*[h]{\scriptsize Structure complete}){$b$ contains EOF}{
        \lIf{$B$ is stable or max rollbacks exceeded}{\Return{$B$}}
        \While(\tcp*[h]{\scriptsize Rollback if unstable}){$B$ is unstable}{
            $ \mathcal I \gets$ indices of unstable bricks in $B$\;
            $i \gets \min \mathcal I$\tcp*[l]{\scriptsize   idx of 1st unstable brick}
            $B \gets [b_1,\dots, b_{i -1}]$\;
        }
    }
}

\end{algorithm}

\input{figTex/results}

\subsection{Brick Texturing and Coloring}
\lblsec{texturing}

While we primarily focus on generating the \textit{shape} of a brick structure,  color and texture play a critical role in creative designs. 
Therefore, we propose a method that applies detailed UV textures or assigns uniform colors to individual bricks. %

\myparagraph{UV Texture Generation.} Given a structure $B$ and its corresponding mesh $\mesh$, we first identify the set of occluded bricks $B_{\text{occ}}$ that have all six faces covered by adjacent bricks, and remove $B_{\text{occ}}$ for efficiency. The remaining bricks $B_{\text{vis}} = B \setminus B_{\text{occ}}$ are merged into a single mesh $\mesh$ with cleaned overlapping vertices using \textit{ImportLDraw}~\cite{ImportLDraw}. We generate a UV map $\text{UV}_{\mesh}$ by cube projection. The texture map $I_{\text{texture}}$ is then generated using FlashTex~\cite{deng2024flashtex}, a fast text-based mesh texturing approach:
\begin{equation}
I_{\text{texture}} = \text{FlashTex}(\mesh, \text{UV}_{\mesh}, \texturecaption),
\end{equation}
where text prompt $\texturecaption$ describes the visual appearance. This texture can be applied through UV printing or stickers. %

\myparagraph{Uniform Brick Color Assignment.} We can also assign each brick a uniform color from a standard color library~\cite{ldraw}. Given a structure $B$, we convert it to a voxel grid $\voxelgrid$ and then to a UV-unwrapped mesh $\mesh_{\voxelgrid}$.
For every voxel $v \in \voxelgrid$, let $f^v_i, i=1,\dots,N_v$ be its visible faces where $0 \le N_v \le 6$.
Each face $f^v_i$ is split into two triangles and mapped to a UV region $\mathcal{S}^v_i$, creating a mesh $\mesh_{\voxelgrid}$ with UV map $\text{UV}_{\voxelgrid}$. We apply FlashTex~\cite{deng2024flashtex} to generate a texture $I_{\text{texture}}$: %

\begin{equation}
    I_{\text{texture}} = \text{FlashTex}(\mesh_{\voxelgrid}, \text{UV}_{\voxelgrid}, \texturecaption). 
\end{equation}
The color of each voxel $\mathcal{C}(v) \in \mathbb{R}^3$ is computed as:
\begin{equation}
    \mathcal{C}(v) = \frac{1}{N_v} \sum_{i=1}^{N_v} \mathcal{C}(f^v_i), \quad \forall v \in \voxelgrid,
\end{equation}
where $\mathcal{C}(f^v_i)=  \frac{1}{|\mathcal{S}^v_i|} \sum_{(x,y) \in \mathcal{S}^v_i} I_{\text{texture}}(x,y)$ is the color of each visible face $f^v_i$, and $|\mathcal{S}^v_i|$ represents the number of pixels in region $\mathcal{S}^v_i$ in the UV map. For each brick $b_t$ and its constituent voxels $\mathcal{V}_t$, we compute the brick color $ \mathcal{C}(b_t) = \frac{1}{|\mathcal{V}_t|} \sum_{v \in \mathcal{V}_t} \mathcal{C}(v)$. 
Finally, we find the closest color in the color set. 
While UV texturing offers higher-fidelity details, uniform coloring allows us to use standard toy bricks.

%% file: figTex/force_illustration.tex
\begin{figure}
    \centering
    \includegraphics[width=0.90\linewidth]{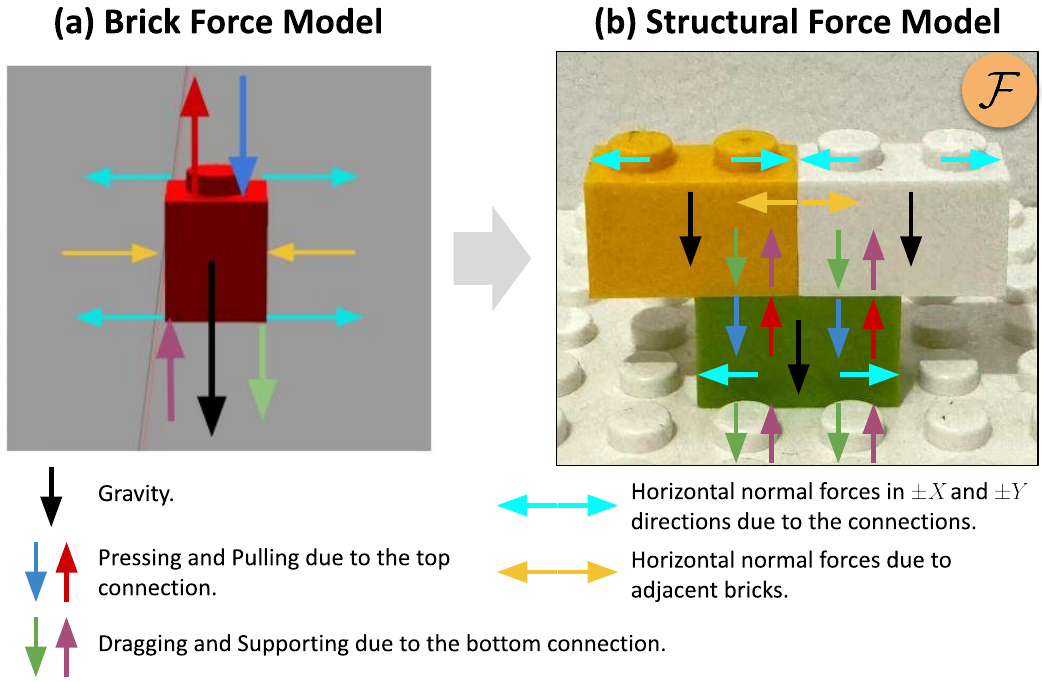}
    \caption{\textbf{Force Model.} (a) We consider all forces exerted on a single brick, including gravity (black), vertical forces with the top brick (red/blue) and bottom brick (green/purple), and horizontal (shear) forces due to knob connections (cyan), and adjacent bricks (yellow). (b) The structural force model $\mathcal{F}$ extends the individual force model to multiple bricks. Solving for static equilibrium in $\mathcal{F}$ determines each brick's stability score.}
    \lblfig{force_model}
    \vspace{-7pt}
\end{figure}

%% file: figTex/results.tex
\begin{figure*}[htpb!]
\begin{minipage}{\textwidth}
        \centering
        \captionof{table}{
        \textbf{Quantitative Results.} We evaluate our method against several baselines on validity (no out-of-library, out-of-bounds, or colliding bricks), stability, CLIP-based text similarity, and DINOv2-based image similarity. Stability, CLIP, and DINO are computed over valid structures only. For LLaMA-Mesh~\cite{llamamesh}, validity requires a well-formed OBJ file.
        Results marked ``+ our stability analysis'' are augmented by generating multiple structures and choosing the first stable one found (if any).
        Our method outperforms all baselines as well as the ablated setups on validity and stability using our proposed rejection sampling and rollback, while maintaining high text similarity.
        }
        \label{tbl:stability}

        \setlength{\tabcolsep}{5pt}
  \begin{tabular}{lccccccc}
    \toprule
    Method & \% valid & \% stable & mean stability & min stability & CLIP & DINO \\
    \midrule
    Pre-trained LLaMA (0-shot) & 0.0\% & 0.0\% & N/A & N/A & N/A & N/A \\
    In-context learning (5-shot) & 2.4\% & 1.2\% & 0.675 & 0.479 & 0.284 & 0.814 \\
    LLaMA-Mesh~\cite{llamamesh} & 94.8\% & 50.8\% & 0.894 & 0.499 & 0.317 & 0.851 \\
    LGM~\cite{lgm} & \textbf{100\%} & 25.2\% & 0.942 & 0.231 & 0.300 & 0.851 \\
    XCube~\cite{XCube} & \textbf{100\%} & 75.2\% & 0.964 & 0.686 & 0.322 & 0.859 \\
    Hunyuan3D-2~\cite{zhao2025hunyuan3d20scalingdiffusion} & \textbf{100\%} & 75.2\% & 0.973 & 0.704 & \underline{0.324} & 0.868 \\
    \midrule
    LLaMA-Mesh~\cite{llamamesh} + our stability analysis & 94.8\% & 58.0\% & 0.896 & 0.564 & 0.317 & 0.851 \\
    LGM~\cite{lgm} + our stability analysis & \textbf{100\%} & 32.5\% & 0.941 & 0.285 & 0.300 & 0.851 \\
    XCube~\cite{XCube} + our stability analysis & \textbf{100\%} & 83.6\% & 0.963 & 0.754 & 0.322 & 0.859\\
    Hunyuan3D-2~\cite{zhao2025hunyuan3d20scalingdiffusion} + our stability analysis & \textbf{100\%} & 88.4\% & 0.976 & 0.813 & \underline{0.324} & 0.868 \\
    \midrule
    Ours w/o rejection sampling or rollback & 37.2\% & 12.8\% & 0.956 & 0.325 & \textbf{0.329} & \textbf{0.888} \\
    Ours w/o rollback & \textbf{100\%} & 24.0\% & 0.947 & 0.228 & 0.322 & \underline{0.882} \\
    \textbf{Ours (\LegoGPT{})} & \textbf{100\%} & \textbf{98.8\%} & \textbf{0.996} & \textbf{0.915} & \underline{0.324} & 0.880 \\
  \bottomrule
\end{tabular}
    \end{minipage}
    
    \vspace{10pt}
    
\begin{minipage}{\textwidth}
    \centering 
     \includegraphics[width=0.95\linewidth]{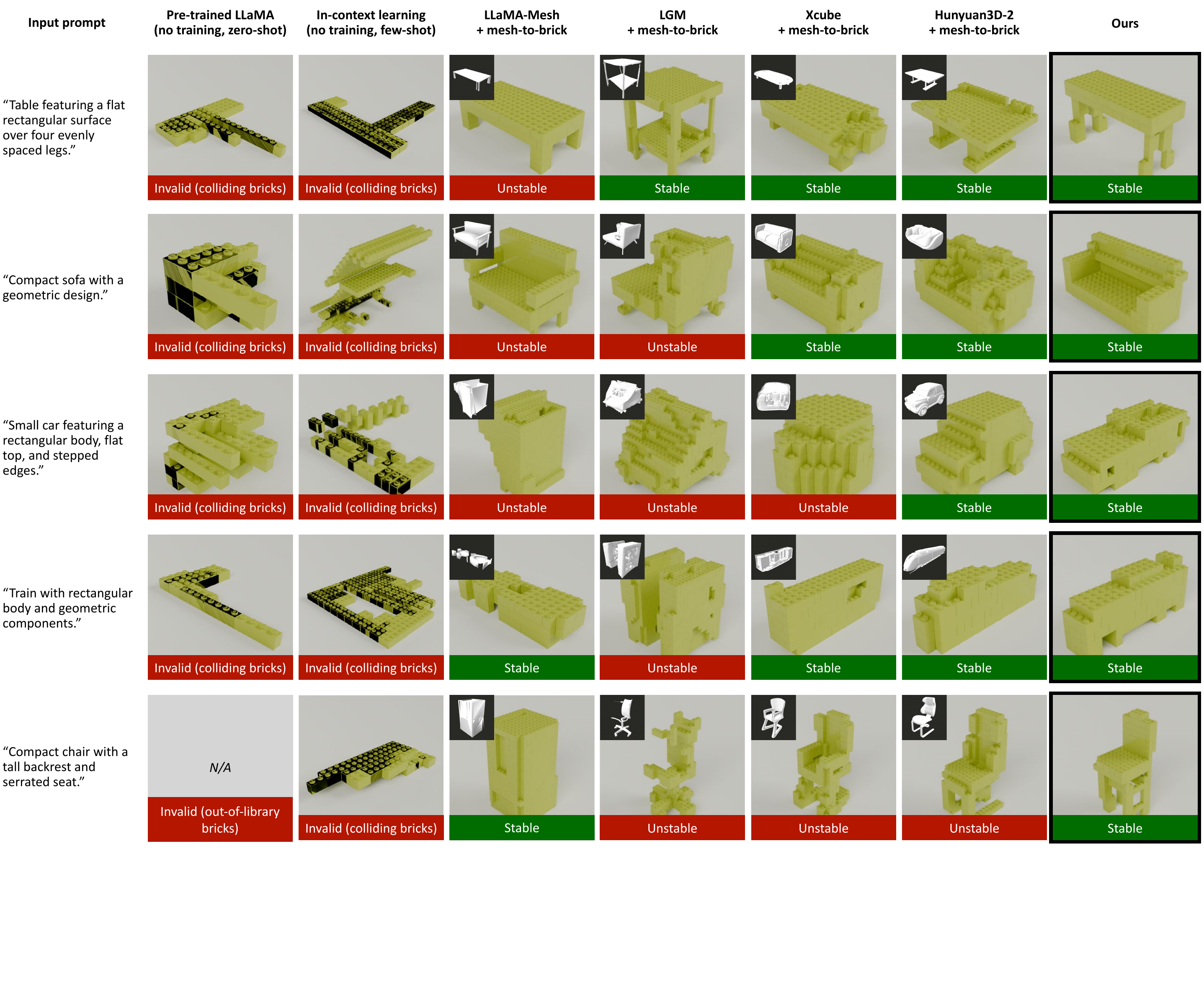}
     \captionof{figure}{\textbf{Result Gallery and Baseline Comparisons.} Our method generates high-quality, diverse, and novel brick structures aligned with the given text prompts. Black bricks are colliding. For LLaMA-Mesh~\cite{llamamesh}, LGM~\cite{lgm}, XCube~\cite{XCube}, and Hunyuan3D-2~\cite{zhao2025hunyuan3d20scalingdiffusion}, an inset of the generated mesh is shown in the top-left corner.}
    \label{fig:results}
    \end{minipage}
\end{figure*}

%% file: sections/04_experiment.tex
\section{Experiments}

\subsection{Implementation Details}

\myparagraph{Fine-tuning.}  
Our fine-tuning dataset contains 240k distinct prompts and 47k+ distinct brick structures. We withhold 10\% of the data for evaluation. For efficiency, we include samples only up to 4096 tokens long. %
Training details are provided in \refapp{sec:training-details}.

\myparagraph{Inference.} To evaluate our method, we generate one brick structure for each of 250 prompts randomly selected from the validation dataset. 
The nonlinear optimization in \refeq{eq:obj} is solved using Gurobi~\cite{gurobi}.
We set $F_T=0.98N$ with $\alpha=10^{-3}$ and $\beta=10^{-6}$.
We allow up to 100 physics-aware rollbacks before accepting the brick structure. The median number of required rollbacks is 2, and the median time to generate one structure is 40.8 seconds.  %

\subsection{Brick Structure Generation Results}
\lblsec{comparison}

\reffig{results} shows a gallery of diverse, high-quality brick structures that closely follow the input prompts.

\input{figTex/ablations}
\input{figTex/texture}

\myparagraph{Baseline Comparisons.} As baselines, we use LLaMA-Mesh~\cite{llamamesh}, LGM~\cite{lgm}, XCube~\cite{XCube}, and Hunyuan3D-2~\cite{zhao2025hunyuan3d20scalingdiffusion} to generate a mesh from each prompt, then convert the meshes to brick structures with our delete-and-rebuild algorithm.
To increase the chance of producing a stable structure, we generate multiple different structures for the same output mesh (10 for LLaMA-Mesh and LGM; 100 for XCube and Hunyuan3D) and accept the first stable one found, if any. Additionally, we compare our method with pre-trained LLaMA models evaluated in both a zero-shot and few-shot manner. For few-shot evaluation, we provide the model with 5 examples of stable brick structures and their captions.

For each method, we compute the proportion of stable and valid structures among the generated designs. Additionally, for each valid structure, we compute its mean and minimum brick stability scores. To evaluate prompt alignment, we compute the CLIP score~\cite{radford2021clip} between a rendered image of each valid structure and the text ``A LEGO model of \{prompt\}''. We also calculate the alignment between rendered images of the generated structure and the ground-truth structure for the same prompt, as measured by the cosine similarity between DINOv2 features~\cite{oquab2024dinov}. As shown in \reftbl{stability}, our method outperforms all baselines in these metrics.

\myparagraph{Ablation Study.} We demonstrate the importance of rejection sampling and physics-aware rollback. As seen in \reffig{ablation}, rejection sampling eliminates invalid bricks, such as those with collisions, while rollback helps to ensure the final assembly structure is stable.
The quantitative results in \reftbl{stability} show that our full method generates a higher proportion of valid and stable brick structures, while closely following the text prompts.

\subsection{Extensions and Applications}
\lblsec{application}

\myparagraph{Robotic Assembly of Generated Structures.} 
We demonstrate automated assembly using a dual-robot-arm system.
The robots use the manipulation policy~\cite{liu2023lightweight} and the asynchronous multi-agent planner~\cite{huang2025apexmr} to manipulate toy bricks and construct the structure.
More details are included in \refapp{sec:robotic_assembly}. 
Since the generated structures are physically stable, efficient automated assembly can be performed.

\myparagraph{Manual Assembly.}
Our generated designs are physically valid and can be hand-assembled. See \refapp{sec:manual_assembly}. %

\myparagraph{Brick Texture and Color Generation.}  \reffig{texture} shows both UV texturing and uniform coloring results of brick structures, demonstrating our method's ability to generate diverse styles while preserving the underlying geometry.

%% file: figTex/ablations.tex
\begin{figure}[t]
    \centering
    \includegraphics[width=0.95\linewidth]{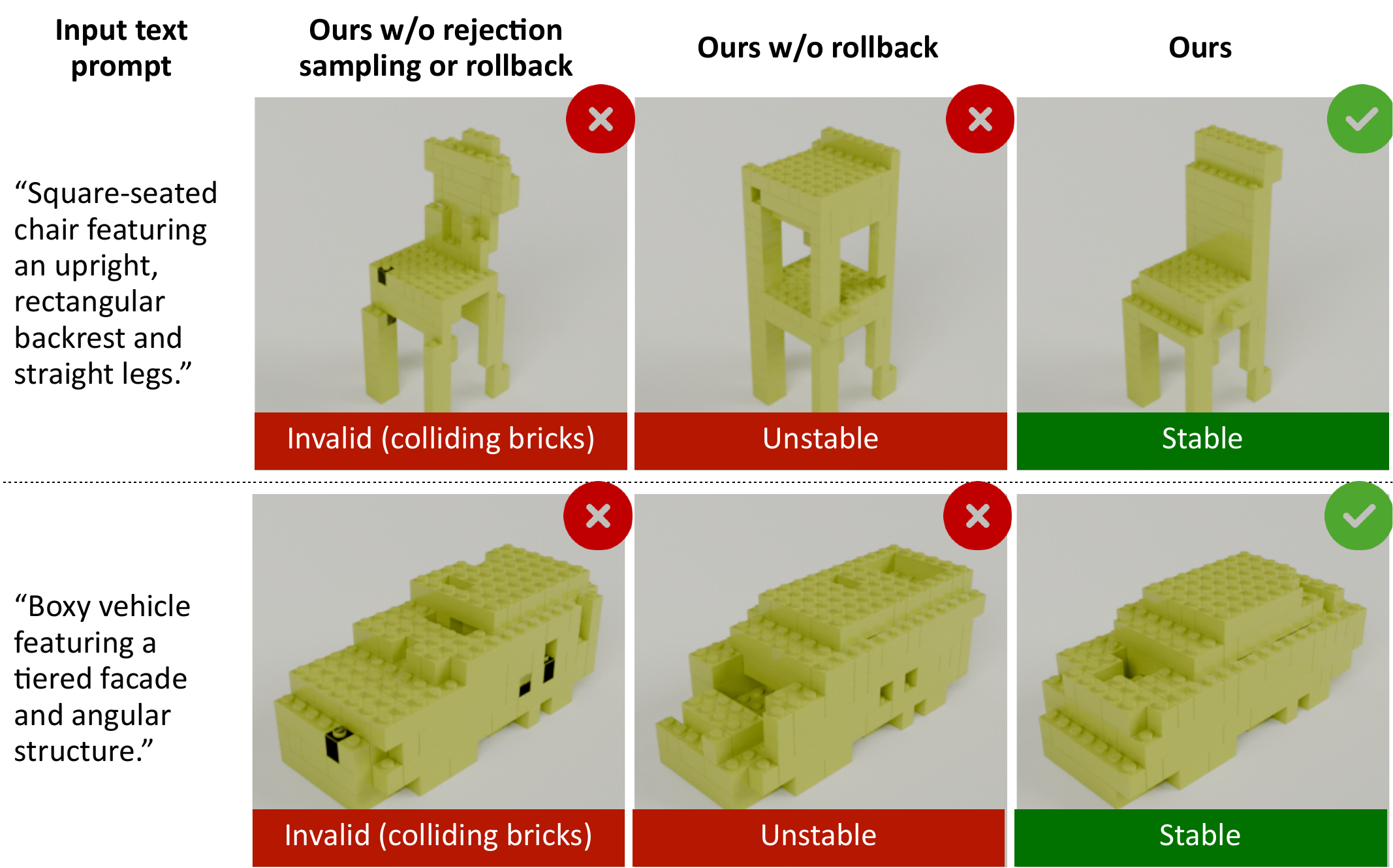}
    \caption{\textbf{Ablation Study.} Brick-by-brick rejection sampling and physics-informed rollback help to ensure that the generated structure is both valid and stable. Black indicates colliding bricks.}
    \label{fig:ablation}
\end{figure}

%% file: figTex/texture.tex
\begin{figure}[t]
    \centering
    \includegraphics[width=0.95\linewidth]{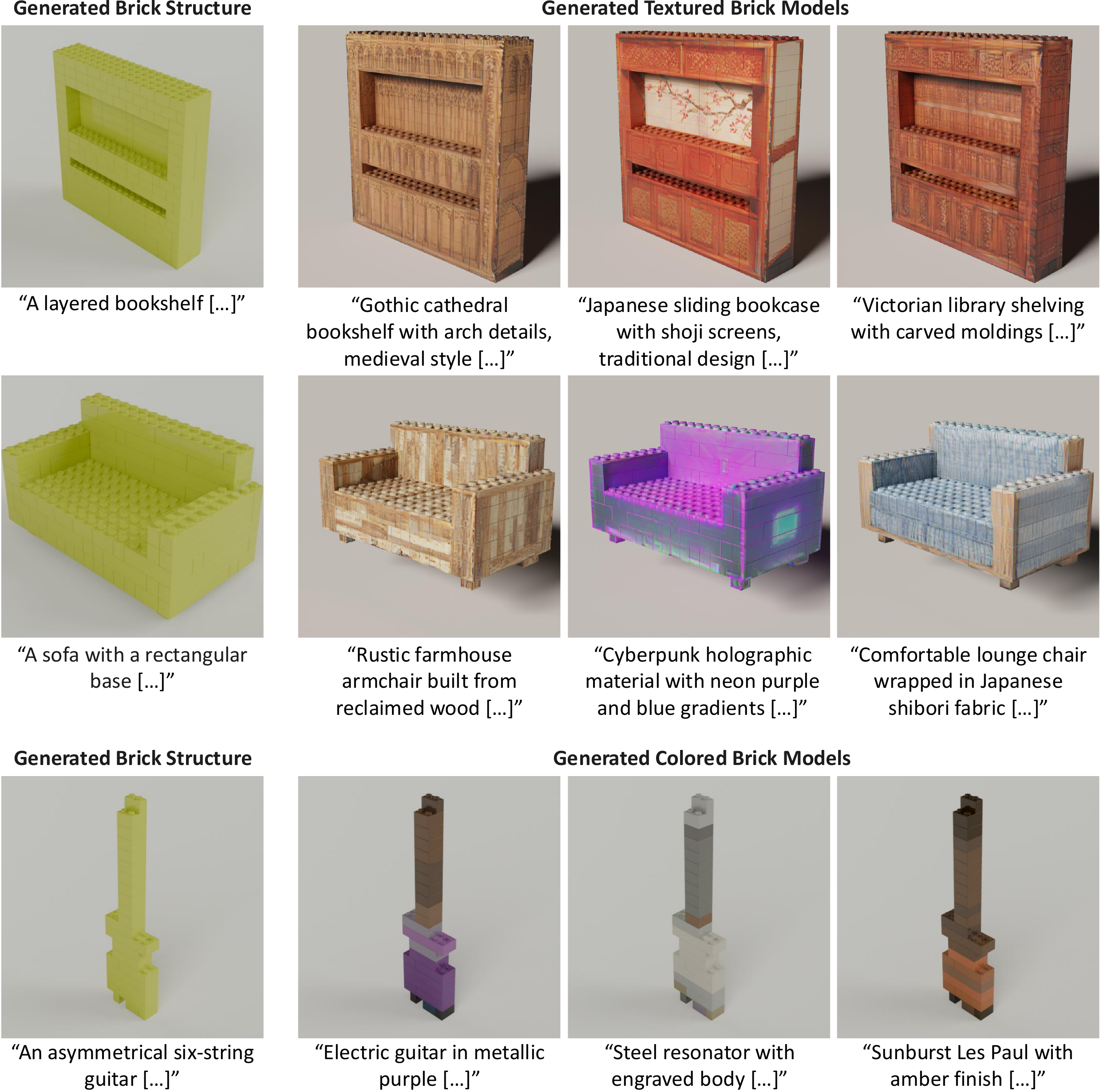}
    \caption{\textbf{Brick Texture and Color Generation.} Our method can generate diverse textured (top two rows) and colored (bottom) structures based on the same shape with different appearance prompts.}
    \lblfig{texture}
    \vspace{-7pt}
\end{figure}

%% file: sections/05_discussion.tex
\section{Conclusion}
\lblsec{discussion}

In this work, we have introduced \LegoGPT{}, an autoregressive model for generating interconnecting toy brick structures from text prompts. Our method predicts the next brick sequentially while ensuring physical stability and buildability. We have shown that our method outperforms LLM backbone models and recent text-to-3D generation methods.

%% file: 06_supplement.tex
\clearpage
\appendix

\section*{Appendix}
\section{\LegoGPT{} Implementation Details} \lblsec{sec:training-details}

\myparagraph{Captioning Details.}
The complete prompt template used for GPT-4o caption generation is as follows:

``This is a rendering of a 3D object built with LEGO bricks with 24 different views. The object belongs to the category of \{CATEGORY\_NAME\}. You will generate five different captions for this \{CATEGORY\_NAME\} that:

    1. Describes the core object/subject and its key geometric features
    
    2. Focuses on structure, geometry, and layout information
    
    3. Uses confident, concrete, and declarative language
    
    4. Omits color and texture information
    
    5. Excludes medium-related terms (model, render, design)
    
    6. Do not describe or reference each view individually.
    
    7. Focus on form over function. Describe the physical appearance of components rather than their purpose. 
    
    8. Describe components in detail, including size, shape, and position relative to other components.
    
    9. The five captions should be from coarse to fine, with the first one being the most coarse-grained (e.g., a general description of the object, within 10 words) and the last one being the most fine-grained (e.g., a detailed description of the object, within 50 words). The five captions should be different from each other. Do not include any ordering numbers (e.g., 1, a, etc.).
    
    10. Describe the object using the category name "\{CATEGORY\_NAME\}" or synonyms of the category name "\{CATEGORY\_NAME\}".''

\myparagraph{\SLV{} Details.}
We generate \SLV{} starting from the objects in ShapeNetCore~\cite{shapenet2015}.
While ShapeNetCore provides both voxel and mesh representations, we find that working directly with mesh data better preserves geometric details.
We voxelize the 3D mesh into a $20\times 20\times 20$ grid representation and generate its brick structure using a delete-and-rebuild algorithm. This algorithm is similar to that studied in prior work~\cite{legolization}. However, instead of initializing the structure by filling its voxels with $1\times 1$ unit bricks and randomly merging them into larger bricks, we greedily place bricks to fill the voxels layer-by-layer from bottom to top. We use eight commonly available standard bricks: $1\times 1$, $1\times 2$, $1\times 4$, $1\times6$, $1\times 8$, $2\times 2$, $2\times 4$, and $2\times 6$. We prioritize placing 1) bricks that are only partially supported by bricks on the layer below, 2) bricks that touch multiple bricks on the layer below, 3) large bricks, and 4) bricks of the opposite orientation from bricks on the layer below.

After initializing the structure, we apply the new stability analysis algorithm \cite{liuStableLegoStabilityAnalysis2024} to iteratively identify weak regions, delete their bricks, and rebuild them by greedily placing bricks, prioritizing 1) bricks that connect multiple disconnected components and 2) large bricks.
This process is stochastic; we choose weak regions randomly as per~\cite{legolization}, and we break ties between two bricks of the same heuristic value randomly.
Using this delete-and-rebuild algorithm, we generate two different structures for each object.

With our delete-and-rebuild shape-to-brick algorithm, we generate $62{,}000+$ brick structures covering the $21$ categories with $31{,}218$ unique 3D objects from ShapeNetCore. 
Among the selected 3D objects, $\sim$92\% (\ie $28{,}822$) of them have at least one stable brick design, which offer $47{,}000+$ stable layouts.
Our \SLV{} dataset significantly expands upon the previous StableLego dataset~\cite{liuStableLegoStabilityAnalysis2024} in several key aspects: it contains $>3\times $ more stable unique 3D objects with $>5\times$ more stable structures compared to $8{,}000+$ in StableLego, spans a diverse set of 21 object categories, and provides detailed geometric descriptions for each shape. 

\input{figTex/one_shot_learning}
\input{figTex/novelty}
\myparagraph{Training Details.} The full prompt that we use to construct our instruction fine-tuning dataset is as follows:

``\textit{[SYSTEM]}You are a helpful assistant.

\textit{[USER]}Create a LEGO model of the input. Format your response as a list of bricks: <brick dimensions> <brick position>, where the brick position is (x,y,z).

Allowed brick dimensions are $1\times 1$, $1\times 2$, $2\times 1$, $1\times 4$, $4\times 1$, $1\times 6$, $6\times 1$, $1\times 8$, $8\times 1$, $2\times 2$, $2\times 4$, $4\times 2$, $2\times 6$, $6\times 2$. 
All bricks are 1 unit tall.

\#\#\# Input:

\textit{[INSERT CAPTION]}''

We fine-tune using low-rank adaptation (LoRA)~\cite{lora} with a rank of 32, alpha of 16, and dropout rate of 0.05. To prevent catastrophic forgetting and training instability, we apply LoRA to only the query and value matrices for a total of 3.4M tunable parameters. We train for three epochs on eight NVIDIA RTX A6000 GPUs. We use the AdamW optimizer~\cite{adamw} with a learning rate of $0.002$, using a cosine scheduler with 100 warmup steps and a global batch size of 64. The total training time is 12 hours.

\myparagraph{Inference Details.} The sampling temperature is set to 0.6 for all experiments. During brick-by-brick rejection sampling, to mitigate repeated generation of rejected bricks, we increase the temperature by 0.01 each time the model generates a brick that has already been rejected.

To force each output brick to be in the format ``\{h\}×\{w\} (\{x\},\{y\},\{z\})'', we sample only from the set of valid tokens during each step. For example, the first token must be a digit, the second an `×', and so on. This has little to no effect on the output of \LegoGPT{}, but helps force the baseline pre-trained models to output bricks in the specified format.

In our experiments, only 1.2\% of the generated designs exceed the maximum number of rollbacks and fail to produce a stable final structure.

\myparagraph{Novelty Analysis.} For each generated structure, we find its closest structure in the training dataset, measured by computing the Chamfer distance in voxel space. As seen in \reffig{novelty}, the generated structures are distinct from their nearest neighbors in the dataset, confirming that our model can create novel designs rather than simply memorizing the training data.

\myparagraph{In-context Learning.} We use LLaMA-3.2-1B-Instruct~\cite{dubey2024llama} as our base model, chosen for its computational efficiency. 
The in-context learning pipeline is shown in \reffig{llama-one-shot-learning}, where the base model can generate brick structures through in-context learning, while suffering from collisions and disconnectivity. We do not use rejection sampling or rollback when evaluating zero-shot or few-shot generation, as doing so results in an excessive number of rejections and a sharp increase in generation time.

\input{figTex/robot_execution}

\section{Robotic Assembly} \lblsec{sec:robotic_assembly}

We demonstrate automated assembly using a dual-robot-arm system as shown in \reffig{robot_execution}.
The system consists of two Yaskawa GP4 robots, each equipped with an ATI force-torque sensor.
A calibrated baseplate is placed between them, and the robots use the bricks initially placed on the plate to construct the brick structure.
Given a brick structure $\legostructure$, we employ the action mask in \cite{liu2024physics} with assembly-by-disassembly search \cite{tian2022assemble} to generate a physically executable assembly sequence for the robots, \ie reordering the brick sequence so that 1) each intermediate structure is physically stable by itself, 2) each intermediate structure is stable under the robot operation, and 3) each assembly step is executable within the system's dexterity.
With the executable assembly sequence, an asynchronous planner \cite{huang2025apexmr} distributes the assembly tasks to the robots, plans the robots' movements, and coordinates the bimanual system to construct the brick structure.
The robots use the end-of-arm tool and the manipulation policy presented in \cite{liu2023lightweight,huang2025apexmr} with closed-loop force control to robustly manipulate bricks and construct the structure.

\section{Manual Assembly}\lblsec{sec:manual_assembly}
As shown in \reffig{fig:manual_assembly}, human users can assemble our generated structures, demonstrating their physical stability and buildability. 
Notably, since our method outputs a sequence of intermediate steps, it naturally serves as an intuitive assembly guide.

\input{figTex/manual_assembly}

\section{Limitations} \lblsec{limitations} 

Though our method outperforms existing methods, it still has several limitations. 
First, due to limited computational resources, we have not explored the largest 3D dataset. As a result, 
our method is restricted to producing designs within a $20\times 20\times 20$ grid across 21 categories, while recent 3D generation methods can create a wider variety of objects. Future work includes scaling up model training at higher grid resolutions on larger, more diverse datasets, such as Objaverse-XL~\cite{objaversexl}. Training on large-scale datasets can also improve generalization to out-of-distribution text prompts. 

Second, our method currently supports a fixed set of commonly used toy bricks. In future work, we plan to expand the brick library to include a broader range of dimensions and brick types, such as slopes and tiles, allowing for more diverse and intricate designs.

%% file: figTex/one_shot_learning.tex
\begin{figure}
    \centering
    \includegraphics[width=\linewidth]{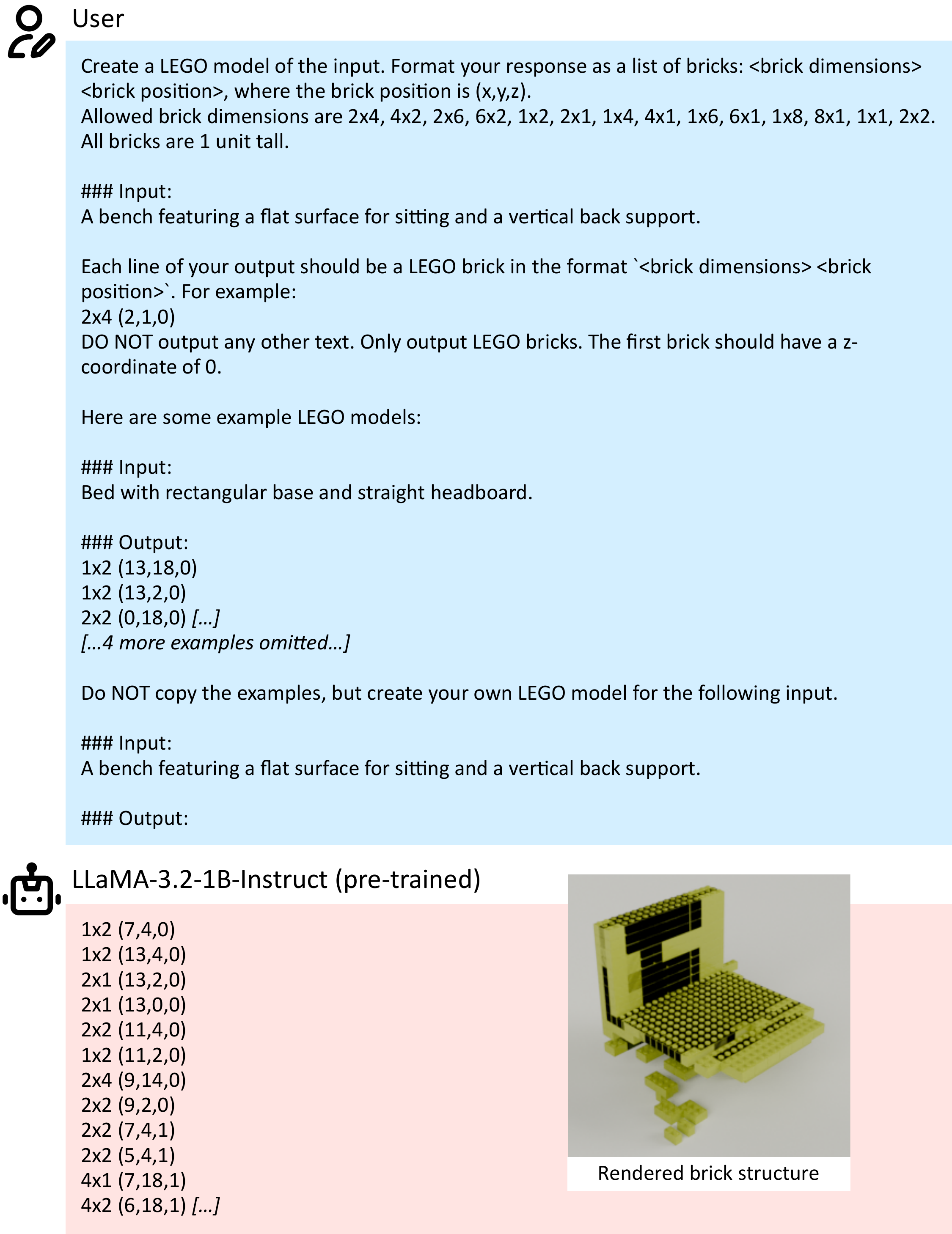}
    \caption{\textbf{Few-shot Learning.} Given a prompt and several example structures in text format, a pre-trained LLaMA model can generate brick designs with some structure.}
    \label{fig:llama-one-shot-learning}
    \vspace{-15pt}
\end{figure}

%% file: figTex/novelty.tex
\begin{figure*}[t]
    \centering
    \includegraphics[width=\linewidth]{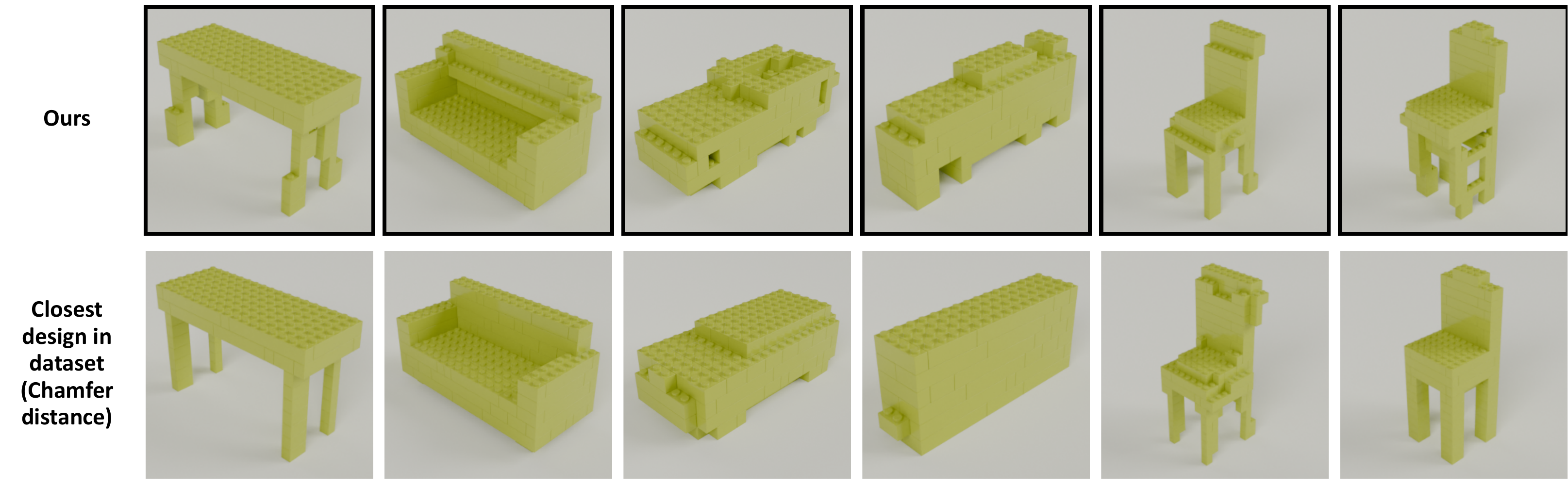}
    \caption{\textbf{Novelty Analysis.} For each structure generated by \LegoGPT{}, we find the closest structure in the training dataset as measured by Chamfer distance in voxel space. The generated structures are distinct from their nearest neighbors, indicating low memorization.}
    \label{fig:novelty}
    \vspace{-15pt}
\end{figure*}

%% file: figTex/robot_execution.tex
\begin{figure*}[t]
    \centering
    \includegraphics[width=\linewidth]{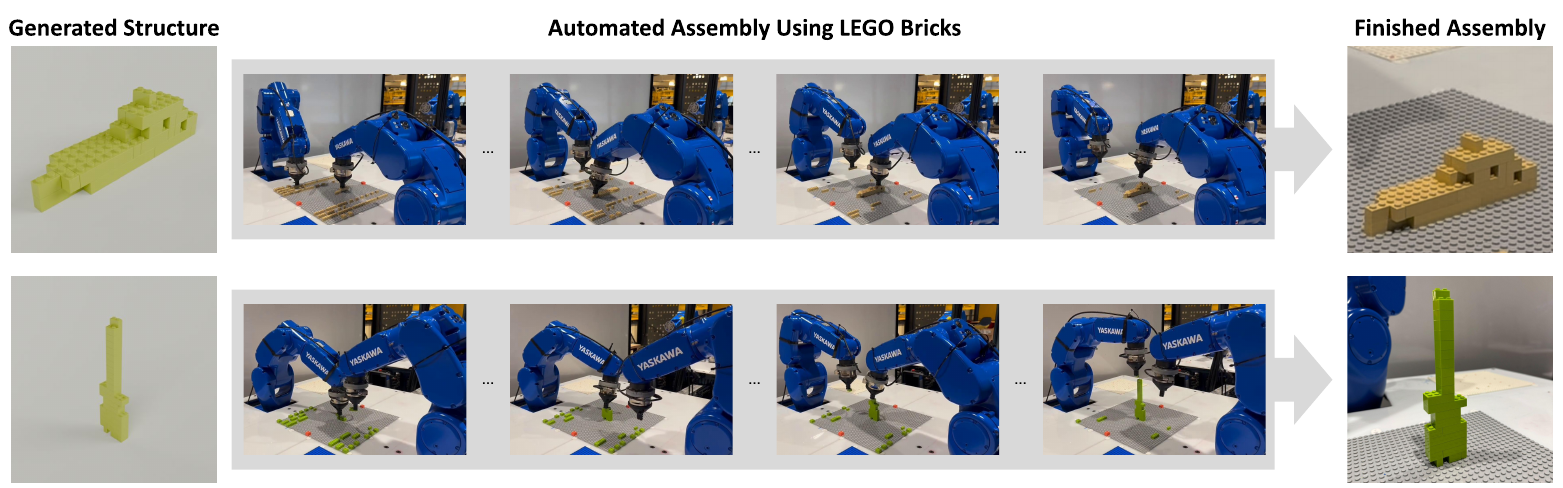}
    \caption{\textbf{Automated Assembly.} We demonstrate robotic assembly of generated structures using LEGO bricks.}
    \lblfig{robot_execution}
    \vspace{-15pt}
\end{figure*}

%% file: figTex/manual_assembly.tex
\begin{figure*}
    \centering
    \includegraphics[width=\linewidth]{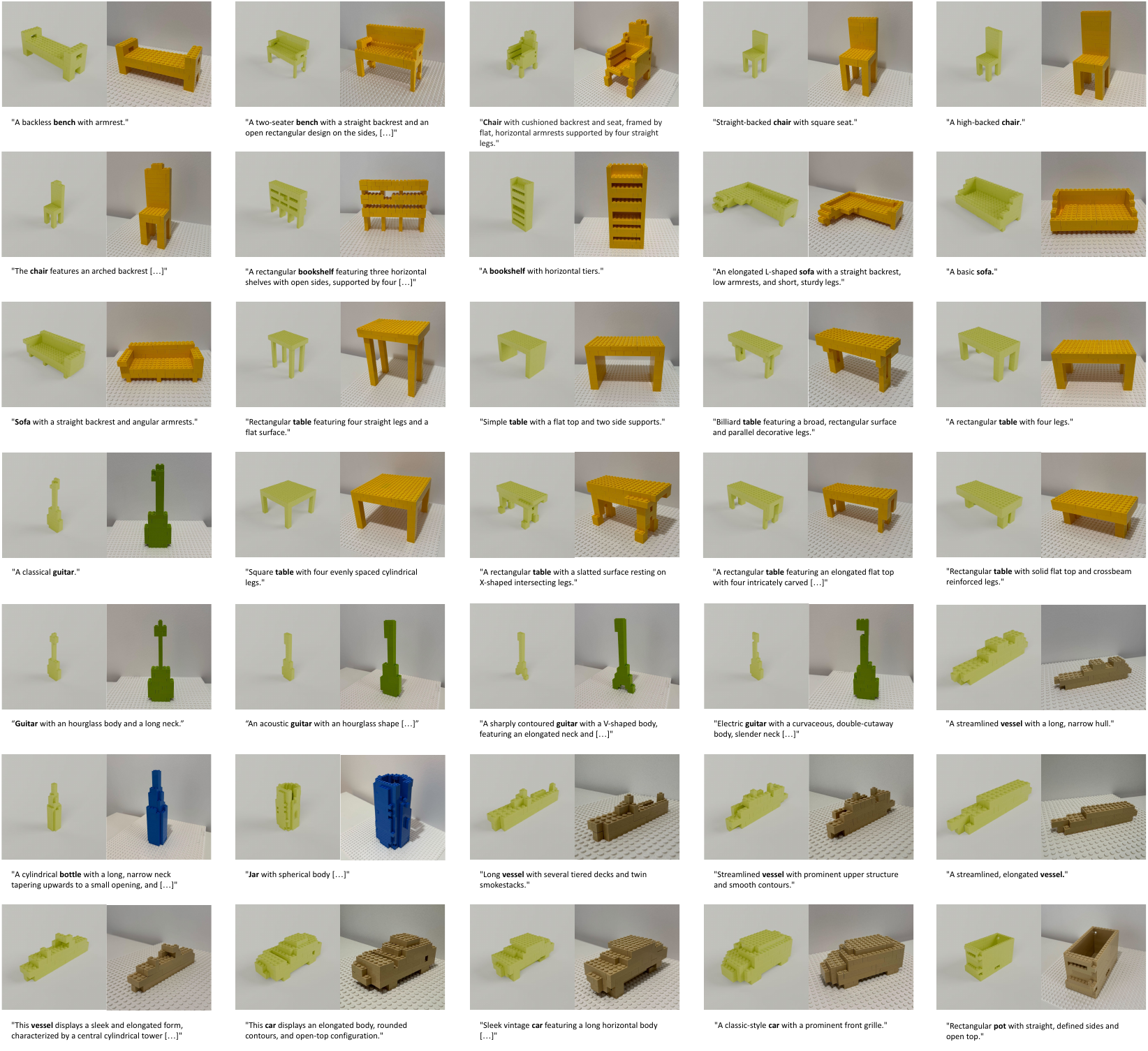}
    \caption{\textbf{Manual Assembly.} We demonstrate manual assembly of generated structures using LEGO bricks.}
    \lblfig{fig:manual_assembly}
\end{figure*}